\documentclass[conference]{IEEEtran}
\IEEEoverridecommandlockouts
\usepackage{cite}
\usepackage{amsmath,amssymb,amsfonts}
\usepackage{algorithmic}
\usepackage{graphicx}
\usepackage{textcomp}
\usepackage{xcolor}
\usepackage{placeins}
\def\BibTeX{{\rm B\kern-.05em{\sc i\kern-.025em b}\kern-.08em
    T\kern-.1667em\lower.7ex\hbox{E}\kern-.125emX}}
\begin{document}

\title{Applying Word Embeddings to Measure Valence in Information Operations Targeting Journalists in Brazil\\

\thanks{This work was generously supported by the Carnegie Endowment for International Peace}
}

\author{\IEEEauthorblockN{David A. Broniatowski}
\IEEEauthorblockA{\textit{Department of Engineering Management and Systems Engineering} \\
\textit{The George Washington University}\\
Washington, DC, USA \\
broniatowski@gwu.edu}}

\maketitle

\begin{abstract}
Among the goals of information operations are to change the overall information environment vis-à-vis specific actors. For example, “trolling campaigns” seek to undermine the credibility of specific public figures, leading others to distrust them and intimidating these figures into silence. To accomplish these aims, information operations frequently make use of “trolls” – malicious online actors who target verbal abuse at these figures. In Brazil, in particular, allies of Brazil’s current president have been accused of operating a “hate cabinet” – a trolling operation that targets journalists who have alleged corruption by this politician and other members of his regime. Leading approaches to detecting harmful speech, such as Google's Perspective API, seek to identify specific messages with harmful content. While this approach is helpful in identifying content to downrank, flag, or remove, it is known to be brittle, and may miss attempts to introduce more subtle biases into the discourse. Here, we aim to develop a measure that might be used to assess how targeted information operations seek to change the overall valence, or appraisal, of specific actors. Preliminary results suggest known campaigns target female journalists more so than male journalists, and that these campaigns may leave detectable traces in overall Twitter discourse. 
\end{abstract}


\section{Introduction}
Online harassment and ``trolling'' campaigns often have political objectives. Among these are attempts to shape public opinion and promote discord, as when the Russian Internet Research Agency attempted to influence the 2016 United States presidential elections \cite{linvill2020troll}. Other campaigns seek to intimidate silence individuals and organizations with expertise or perspectives that could be threatening to a rival political regime, such as when journalists are targeted with hateful rhetoric and threats of harm by political extremists \cite{obermaier2018journalists}. 

\subsection{The Brazilian ``Hate Cabinet''}
In a series of media reports\footnote{https://www1.folha.uol.com.br/folha-topicos/gabinete-do-odio/} Brazilian President Jair Bolsonaro has been accused of of operating a ``Hate Cabinet'' -- a large network of online ``trolls'' that are charged with harassing journalists who report negatively on the Bolsonaro regime. Although Bolsonaro has denied these claims, several journalists have reported being the target of these attacks\footnote{https://latamjournalismreview.org/articles/women-journalists-receive-more-than-twice-as-many-insults-on-twitter-than-male-colleagues/} with women and members of underrepresented minorities especially likely to have been targeted.\footnote{https://rsf.org/en/news/brazil-quarterly-analysis-president-bolsonaros-systematic-attempts-reduce-media-silence} 
These campaigns and other influence operations intend to induce  changes in public opinion regarding controversial issues, while silencing dissenting voices, such as those of journalists, who promote narratives that are perceived as hostile to the interests of the regime. 
In order to know how best to respond to these organized campaigns, we must first measure their efficacy. Therefore, we seek to develop a novel natural language processing tool that might be used to detect these online harassment campaigns. Specifically, we draw upon automated techniques  that are designed to measure bias and valence – proxies for broad associations with public opinion – in online corpora.  

\section{Literature Review}
Our analysis is based upon the popular Word2Vec model \cite{mikolov2013efficient} – an algorithim that has been widely used to generate ``word embeddings'' semantic spaces in which words with similar meanings are collocated in a high-dimensional vector space. Importantly for our application, Word2Vec also captures differences in these semantics that are interpretable. For example, in a classic example, the vector representing the word “queen” is close to the vector obtained from the following expression: 
\begin{equation}
    queen \approx king – man + woman
\end{equation}
Similar results are obtained when examining the relationships between nations and their capitol cities; for example,
\begin{equation}
    Berlin \approx Paris – France + Germany
\end{equation}

In prior work \cite{toney2021automatically}, we found that similar word embedding models could be used to characterize information operations that were targeted at presidential candidates during the US presidential elections of 2016. Specifically, a known “trolling campaign” supporting candidate Donald Trump, and opposing candidate Hilary Clinton was run by the Russian Internet Research Agency (IRA). Using tweets from this campaign, we found that words referencing candidate Trump were more closely related to words indexing trust whereas words referencing candidate Clinton were more closely related to words indexing distrust \cite{toney2021automatically}. Similarly, early during the outbreak of the COVID-19 pandemic, we found that tweets containing anti-Chinese hashtags significantly associated Russia with words indicating calm and pleasantness \cite{toney2021automatically}. This approach is unique because its unit of analysis is the overall corpus, not just the individual tweet. Thus, we aim to use this technique to determine if a corpus of tweets expresses an aggregate position regarding a specific target in an information operation. 

\section{Methodology}
\subsection{Synchronic Analysis: Validation Study}
\subsubsection{Dataset}
Using the Social Feed Manager software \cite{Littman_2020}, we collected data from Twitter’s streaming API containing the hashtag \#GloboLixo between June 2 and September 4, 2021. This query returned 355,068 tweets. We used this hashtag, which roughly translates as “Globo is Trash” because it expresses disgust with the Brazilian media company Globo News, a prominent Brazilian news source that has criticized the government, and the journalists who work for it. In particular, this hashtag is widely used by Twitter accounts supporting Brazilian president Jair Bolsonaro, which have allegedly been used to attack journalists who report on alleged corruption by the Bolsonaro regime. 
\subsubsection{Choice of Algorithm}
Given this dataset, our next steps was to project the corresponding corpus into a word embedding space. To do so, we relied upon pretrained language models. Specifically, since the tweets in our dataset were written in Brazilian Portuguese, we downloaded several pretrained Brazilian Portuguese corpora \cite{hartmann2017portuguese}, including those trained using two different variations of the Word2Vec algorithm: a) Continuous Bag-of-Words (CBOW), and b) Skip-Gram with Negative Sampling (SGNS), and GloVe \cite{pennington2014glove}. All corpora used 300 dimensions. 
\subsubsection{Choice of Constructs}
We fine-tuned these pre-trained corpora using the Word2Vec algorithm trained on the \#GloboLixo dataset, using both the CBOW and SGNS methods. We also experimented with fixing the locations of the word vectors corresponding All Word2Vec models were implemented using the Gensim python package with default settings \cite{rehurek2011gensim}. Since these algorithms are stochastic in nature, we fit 10 Word2Vec models for each combination of algorithm and pre-trained corpus.
After having trained these language models, we used a variant of the Word Embedding Association Test (WEAT; \cite{caliskan2017semantics}) to generate scores reflecting the valence and trustworthiness of a set of journalists that were identified by domain experts (in order to protect journalists' identities, this available upon request from the author for legitimate research purposes). The WEAT operates on two sets of target words (e.g., words representing male journalists
and words representing female journalists) and two sets
of polar attributes (e.g., words representing pleasantness and words representing unpleasantness) and computes an effect size (Cohen’s d) to measure the bias associations between the target sets and polar attribute sets. Our variation of the WEAT returns a value for a specific journalist given a set of polar attributes. Follow \cite{caliskan2017semantics}, let X and Y be two target word sets of equal size and a be the Twitter handle of a journalist. The score for that journalist given those target words is given by the following formula:

\begin{equation}
s(X,Y,a) = \frac{\frac{1}{m}(\sum_{\Vec{x} \in X} cos(\Vec{a},X)-\sum_{\Vec{y} \in Y} cos(\Vec{a},Y))}{ \sigma_{\Vec{w} \in X \cup Y} cos(\Vec{a},w)}
\end{equation}

where $\sigma$ denotes standard deviation and similarity between two vectors in a word embedding space is calculated using cosine similarity. We calculated these scores for 58 journalists whose Twitter handles appeared in the \#GloboLixo dataset. Following Toney et al. \cite{toney2020valnorm}, we used words corresponding to positive and negative valence. In addition, since these attacks may be associated with undermining trust in specific journalists' professional integrity, we used words associated with trust and distrust \cite{caliskan2017semantics}. Finally, since we observed that several tweets targeting journalists expressed moral opprobrium, and especially disgust, we calculated a “purity” score for each journalist using a similar technique. The specific words used to calculate purity were derived from the Brazilian Portuguese Moral Foundations Dictionary for Fake News classification \cite{carvalho2020brazilian}. We retained all words from this dictionary if they were present in our pretrained Brazilian Portuguese corpora \cite{hartmann2017portuguese}. In each case, words were selected based upon their presence in these word lists, with words only excluded if they did not appear in our pretrained corpus or the corresponding tweets. When such words were not present, we ensured balance between positive and negative word lists by selecting a subset of words from each list uniformly at random. We found that our scores were robust to specific word selection (for any given model, Cronbach’s $\alpha$ values were $>$0.95 for 10 different random word selections).  Finally, for comparison purposes, we also included a “null” list of words selected uniformly at random from the Brazilian Portuguese corpus and a “saturated” set of the top eight words from the union of each lists that were visually most- and least- similar to a set of Globo journalists who were known to be targeted in a TSNE plot containing these journalists, all words in the preceding word lists, and a list of abusive words selected by domain experts from the \#GloboLixo corpus (see Appendix A). The intent in including the null and saturated wordlists was to compare the valence, trust, and purity lists to underfit and overfit models, respectively.  The specific words  used in each attribute set were:
\begin{enumerate}
    \item Negative valence vs. positive valence \cite{toney2020valnorm}: \begin{enumerate}
        \item Negative valence: \textit{assalto, assassinato, acidente, agonia, cadeia, cancro, colisão, desastre, divórcio,  enfermidade, falha, fedor, feio, ferido, horrível, horroroso, imundície, malvado, matar, mau, maus-tratos, morte, ódio, pobreza, podre, poluir, prisão, terrível, tragédia, tristeza, veneno, vômito}
        \item Positive valence: \textit{alegria, alegrar, amanhecer, amigo, amor, arco-íris, carícia, céu, diamante, diploma, família, feliz, férias, gentil, glorioso, honesto, honra, leal, liberdade, maravilhoso, milagre, paraíso, paz, prazer, prenda, riso, saúde, sortudo}
    \end{enumerate}
    \item Distrust vs. trust \cite{caliskan2017semantics}: \begin{enumerate}
        \item Distrust: \textit{desleal, desonesto, duvidoso, egoísta, frio, insensível, mesquinho, traiçoeiro, traidor}
        \item Trust: \textit{acolhedor, amigável, amigo, apoiador, bom, confiável, gentil, sincero}
    \end{enumerate}
    \item Impurity vs. purity \cite{carvalho2020brazilian}: \begin{enumerate}
        \item Impurity: \textit{contagiosa, contagioso, corrompe, corrompendo, corromper, corromperam, corrompeu, depravada, depravados, desgraçados, desgraçadamente, doenças, doentes, doentia, doentio, imundice, imundície, imundo, imundos, miseráveis, nojentas, nojentos, pecado, piranha, pródigo, promíscua, puta}
        \item Purity: \textit{abstinência, decência, decente, decentes, igreja, igrejas, incorruptível, inocente, inocentes, integridade, limpa, limpando, limpar, limpas, limpeza, limpo, limpos, piedade, pura, puro, sagrada, sagrado, santa, santana, santo, santos, virgem}
    \end{enumerate}
    \item ``Saturated'' words: \begin{enumerate}
        \item Positive: \textit{acolhedor, bom, confiável, decente, feliz, gentil, honesto, sortudo}
        \item Negative: \textit{corromper, desgraçados, imundos, matar, miseráveis, nojentas, nojentos, puta}
    \end{enumerate}\end{enumerate} 

\subsubsection{Assessing the Validity of Our Measures}
Using the pre-trained Brazilian Corpus, we fit four models trained as follows:\footnote{Although other models are feasible, we did not test them in this work due to time constraints; however, future work may evaluate these}
\begin{enumerate}
    \item A model pretrained using SGNS, with fixed vectors, and fine-tuned using CBOW
    \item A model pretrained using SGNS, with fixed vectors, and fine-tuned using SGNS
    \item A model pretrained using CBOW, with adjustable vectors, and fine-tuned using SGNS
    \item A model pretrained using GloVe, with fixed vectors, and fine-tuned using SGNS
\end{enumerate}
In order to evaluate the reliability of a given model fit, we fit each model 10 times and calculated Cronbach’s alpha between model fits. For each replication of each model, we also computed five traits: trust, purity, valence, null, and saturated scores for each of the 58 journalists in our dataset, averaging across these model fits. These multiple replications allowed us to construct a multi-trait multi-method matrix (MTMM; \cite{campbell1959convergent}) – a classic tool used to evaluate the construct validity of psychometric measures (see Appendix B). This MTMM indicated significant correlations between common traits using Word2Vec embeddings, and between different, but related, traits, using the same embeddings. In particular, as expected, we found that measures of trust, valence, and purity were all significantly correlated with each other, but not correlated with randomly selected words, thus suggesting that these measures possess convergent-discriminant validity. Finally, the MTMM shows that a pre-trained Word2Vec model using SGNS with fixed vectors, and then fine-tuned on Twitter data using CBOW, yielded the largest reliability. We therefore used this model moving forward.
Using this model, we next compared the average valence, trust, and purity scores for each of the 58 journalists in the \#GloboLixo dataset. Specifically, we hypothesized that female journalists in this dataset would have significantly lower trust, purity, and valence scores than male journalists.
\subsection{Diachronic Analysis: Tracking Evolving Information Operations}
\subsubsection{Dataset}
To examine whether we were able to detect these attacks on Twitter, we obtained a list of 204 journalists and their corresponding Twitter handles. We next extracted all tweets containing @mentions of these journalists, between 6/2/2021 and 11/19/2021, yielding roughly 25 million tweets. We next split this dataset into several corpora, each containing all tweets within a given week (e.g., 6/2/2021-6/9/2021; each such corpus contained roughly 1 million tweets). Finally, we fit separate Word2Vec models to each week and calculated the trust, purity, and valence scores for each journalist's Twitter handle. 
We next examined whether these measures might be usable to detect evolving information operations. To do so, we obtained a list of public statements attacking journalists that were made by politicians associated with the Bolsonaro regime. This list, compiled by Brazilian investigate journalists on October 15, 2021, documented 85 such attacks. Of the 204 journalists in our sample, 7 (3.4\%) had been attacked at least once between June and September 2021. We therefore examined how the valence, purity, and trust scores for these journalists changed over time, and especially in the weeks adjacent to the documented attacks. As before, we analyzed female and male journalists separately. 
\section{Results}
\subsection{Synchronic study}
\begin{figure}[htp]

\centering
\includegraphics[width=.45\textwidth]{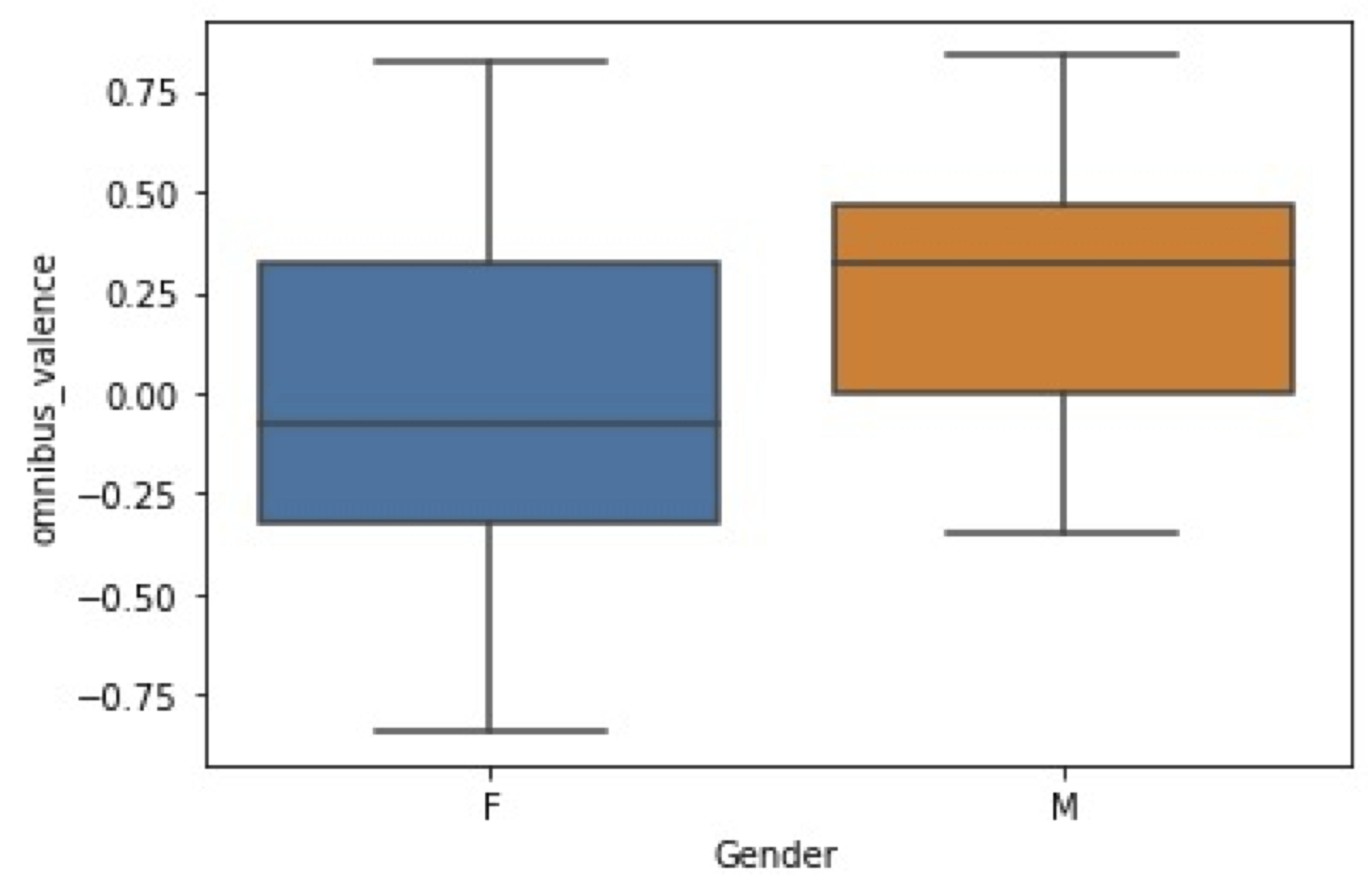}\hfill
\includegraphics[width=.45\textwidth]{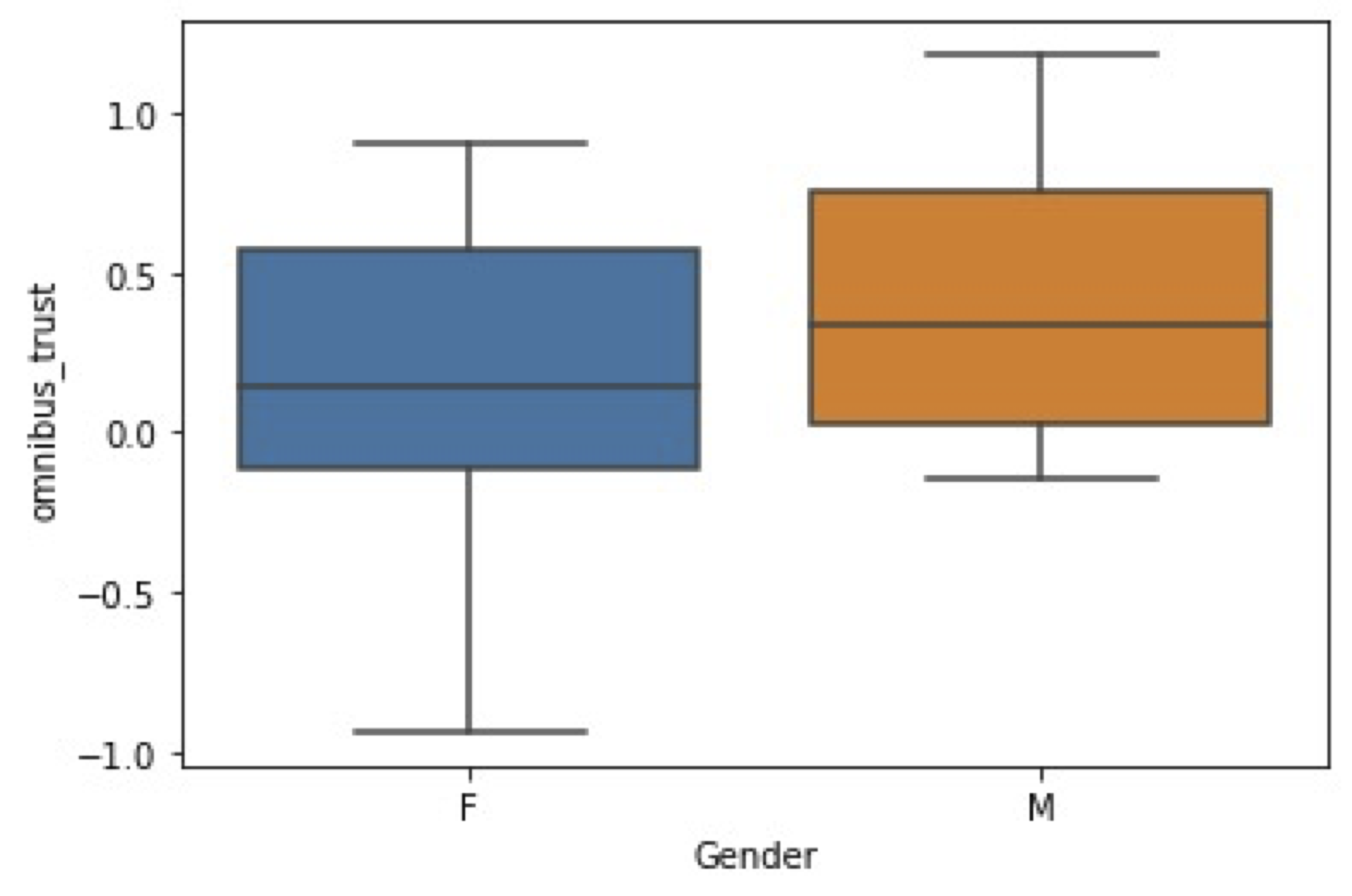}\hfill
\includegraphics[width=.45\textwidth]{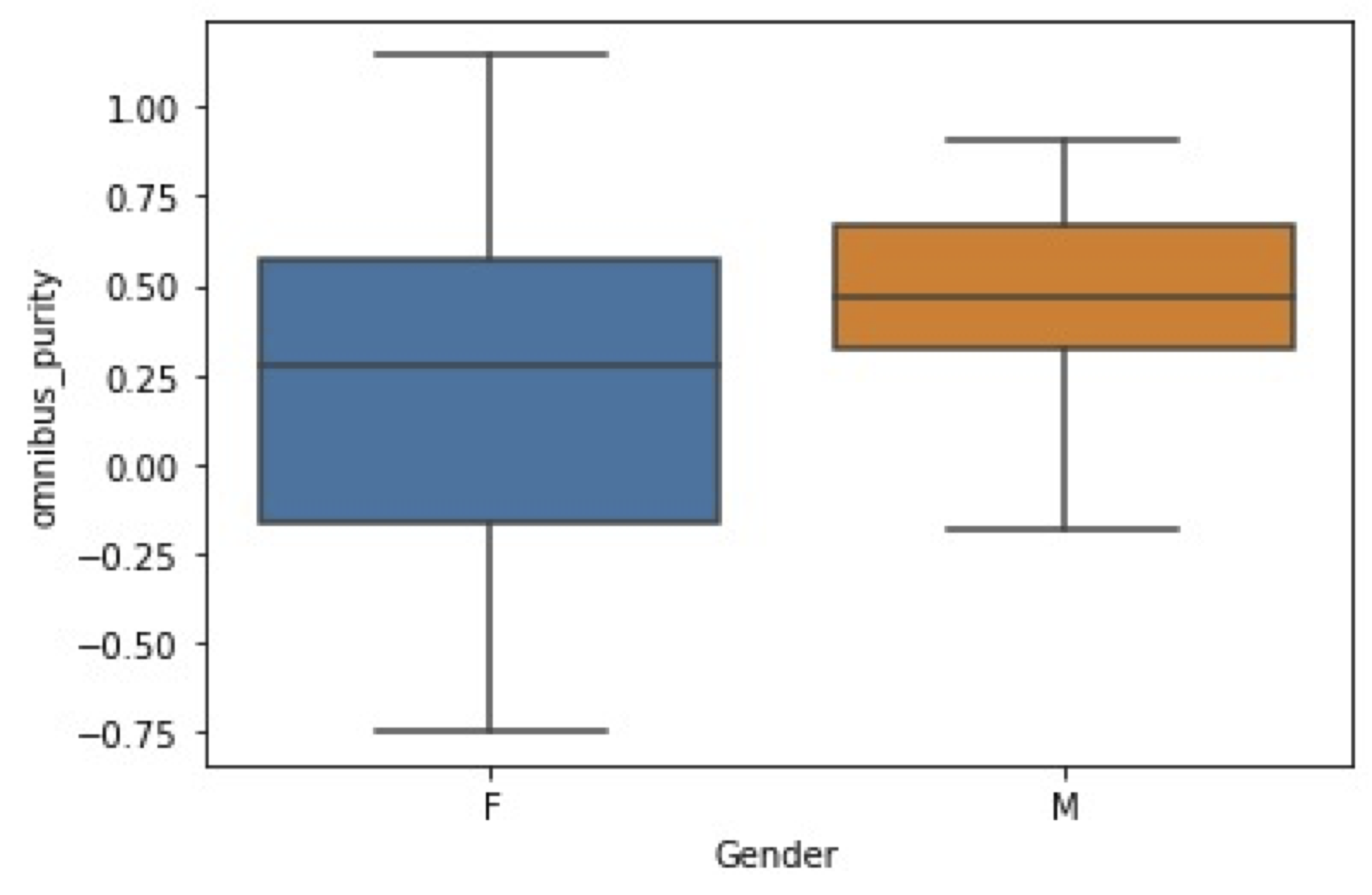}
\caption{Boxplots for valence (top), trust (middle), and purity (bottom) scores for female vs. male journalists in the \#GloboLixo corpus}
\label{fig:boxplots}
\end{figure}

We found that the 37 female journalists in this dataset were framed as significantly less pleasant, \textit{t(56)=2.65, p=0.01}, less trustworthy, \textit{t(56)=2.26, p=0.03}, and less pure, \textit{t(56)=2.41, p=0.02}, than the corresponding 21 male journalists (Figure \ref{fig:boxplots}). 

\begin{figure}[htp]

\centering
\includegraphics[width=.45\textwidth]{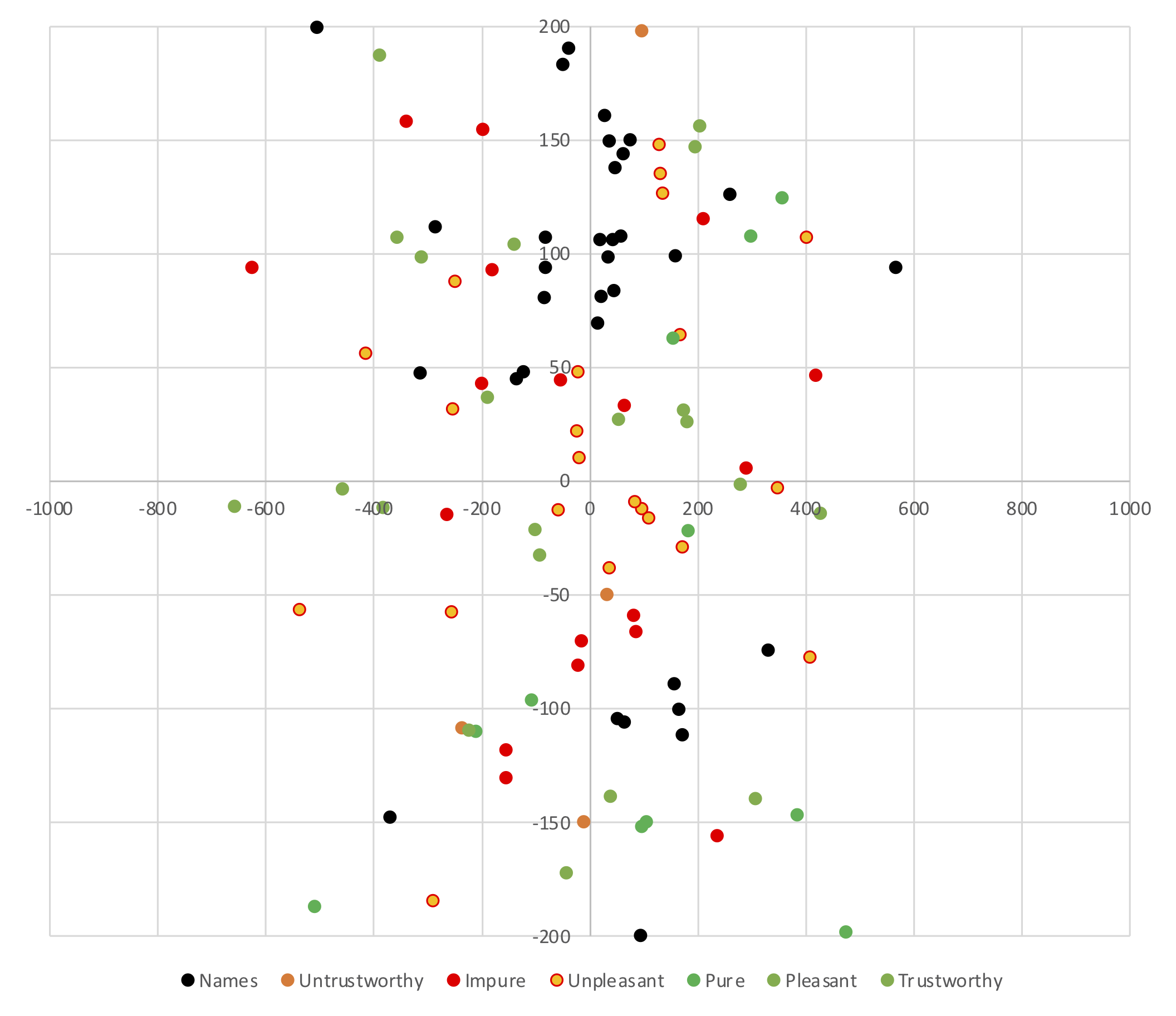}\hfill
\caption{TSNE plot showing clustering of journalist Twitter handles with attribute words.}
\label{fig:TSNE1}
\end{figure}

These results are consistent with a hypothesized gendered nature of these attacks. Furthermore, these results lend an element of predictive validity to our measures.
A visualization of our measures using T-Stochastic Neighbor Embedding (TSNE \cite{van2008visualizing}; Figure \ref{fig:TSNE1}) shows that journalists who were targeted in this campaign fit into two rough clusters associated either with words indicating  unpleasantness or impurity. 

\subsection{Diachronic study}
Of the seven journalists that had been attacked between June and September 2021, four were female and three were male. 
\subsubsection{Journalist F1}
\begin{figure}[htp]

\centering
\includegraphics[width=.45\textwidth]{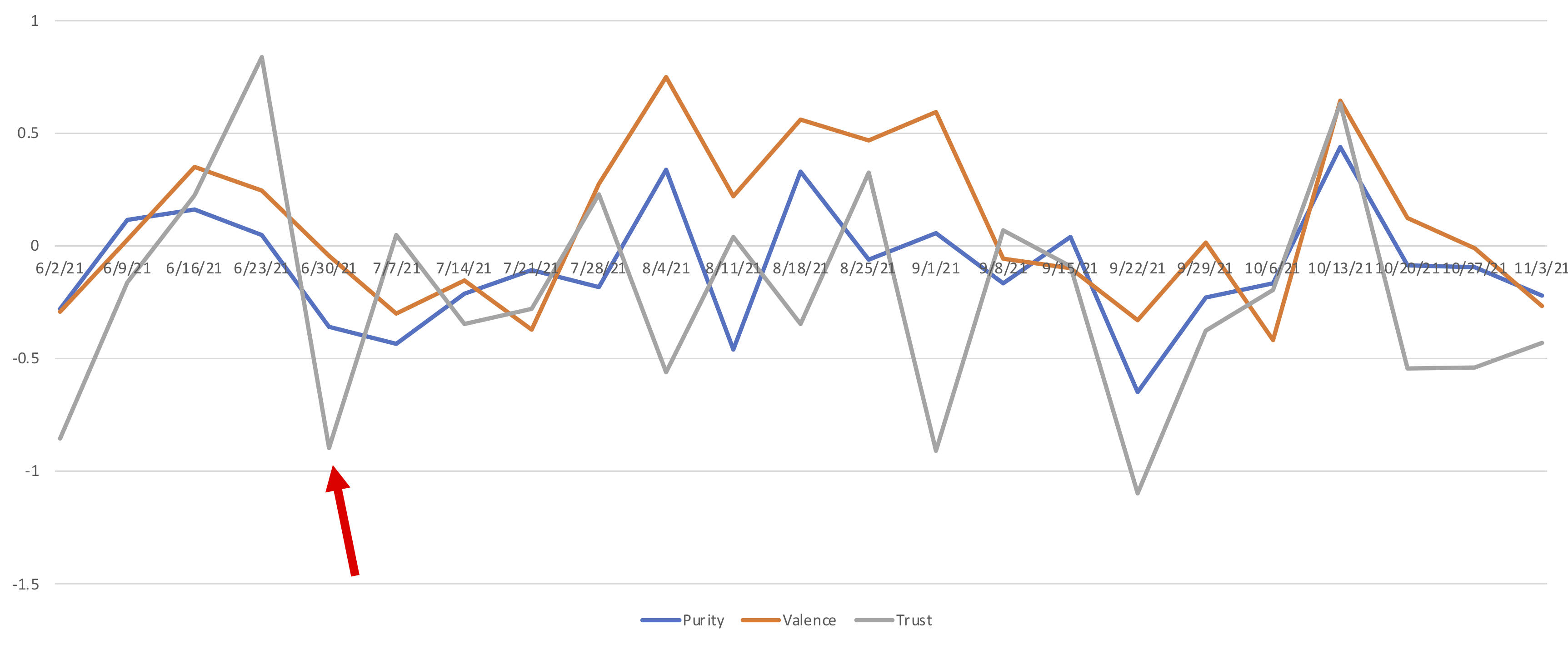}\hfill
\caption{Journalist F1's purity (blue), valence (orange), and trust (grey) scores over time. The red arrow corresponds to the June 29 tweet by Carlos Bolsonaro.}
\label{fig:F1}
\end{figure}

On June 27, 2021, journalist F1 published an interview with a prominent actress who used crude language to accuse the Bolsonaro government of incompetence. Two days later, on June 29, Carlos Bolsonaro, the son of the Brazilian president, posted a tweet linking to this article also using crude language and accusing Journalist F1 of financial corruption. Figure \ref{fig:F1} shows that these events correspond to a drop in Journalist F1's trust score. 

\subsubsection{Journalist F2}
\begin{figure}[htp]

\centering
\includegraphics[width=.45\textwidth]{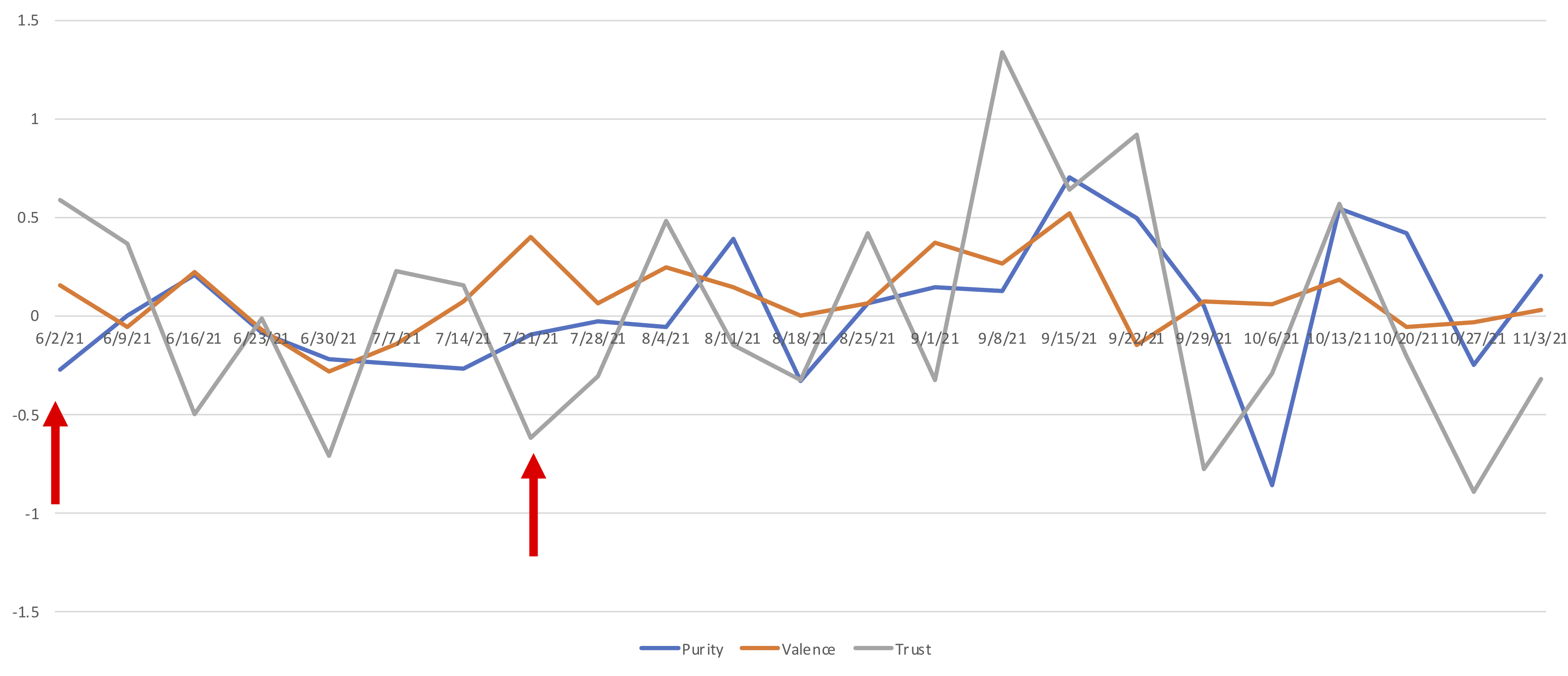}\hfill
\caption{Journalist F2's purity (blue), valence (orange), and trust (grey) scores over time. The red arrows corresponds to the May 27 and July 14 tweets by Eduardo Bolsonaro and President Bolsonaro's advisor, respectively.}
\label{fig:F2}
\end{figure}

Journalist F2 was targeted on two separate occasions: May 27, 2021, and July 14, 2021. In the first attack, Eduardo Bolsonaro, another son of Brazil's president, wrote a series of Twitter posts criticizing the press, which triggered attacks on the social networks of journalist F2. In the second attack, one of President Bolsonaro's advisors accused journalist F2 of a Freudian slip when she referred to President Bolsonaro as ``ex-President'' when reporting upon his evaluation for surgery after the president suffered 10 days of hiccups due to a bowel obstruction. Figure \ref{fig:F2} shows that the first attack correspond to a low point in Journalist F2's purity score and the second attack corresponds to a reduced trust score.

\subsubsection{Journalist F3}
\begin{figure}[htp]

\centering
\includegraphics[width=.45\textwidth]{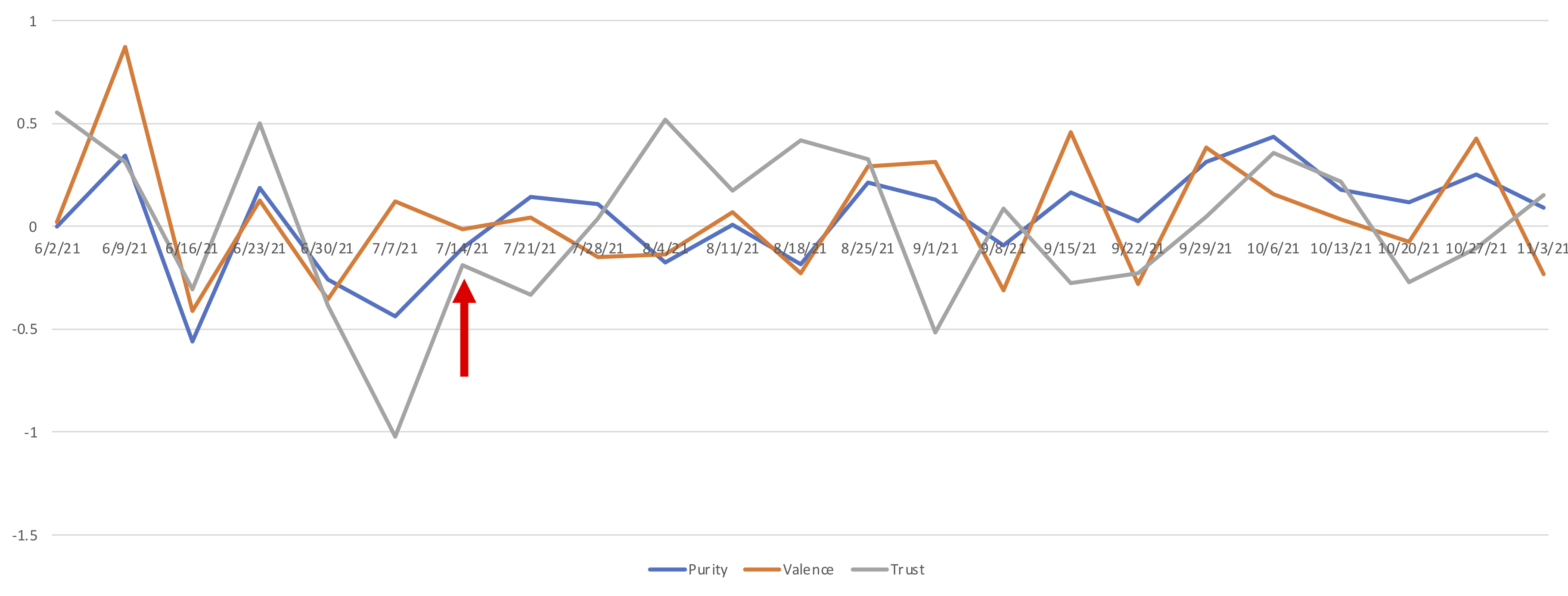}\hfill
\caption{Journalist F3's purity (blue), valence (orange), and trust (grey) scores over time. The red arrow corresponds to the June 30 tweet by Eduardo Bolsonaro.}
\label{fig:F3}
\end{figure}

On June 30, 2021, journalist F3 posted a tweet objecting to President Bolsonaro's religious rhetoric and emphasizing that the Brazilian state is secular. One July 13, Eduardo Bolsonaro responded by calling for ``spiritual war'' and accusing journalist F3 of being a ``leftist militant''. Figure \ref{fig:F3} shows that journalist F3's trust score declined precipitously following following the Twitter post on June 30th, but that this trust score was increasing preceding Eduardo Bolsonaro's tweet. However, in the week immediately following the tweet, journalist F3's trust score declined once again. 

\subsubsection{Journalist F4}
\begin{figure}[htp]

\centering
\includegraphics[width=.45\textwidth]{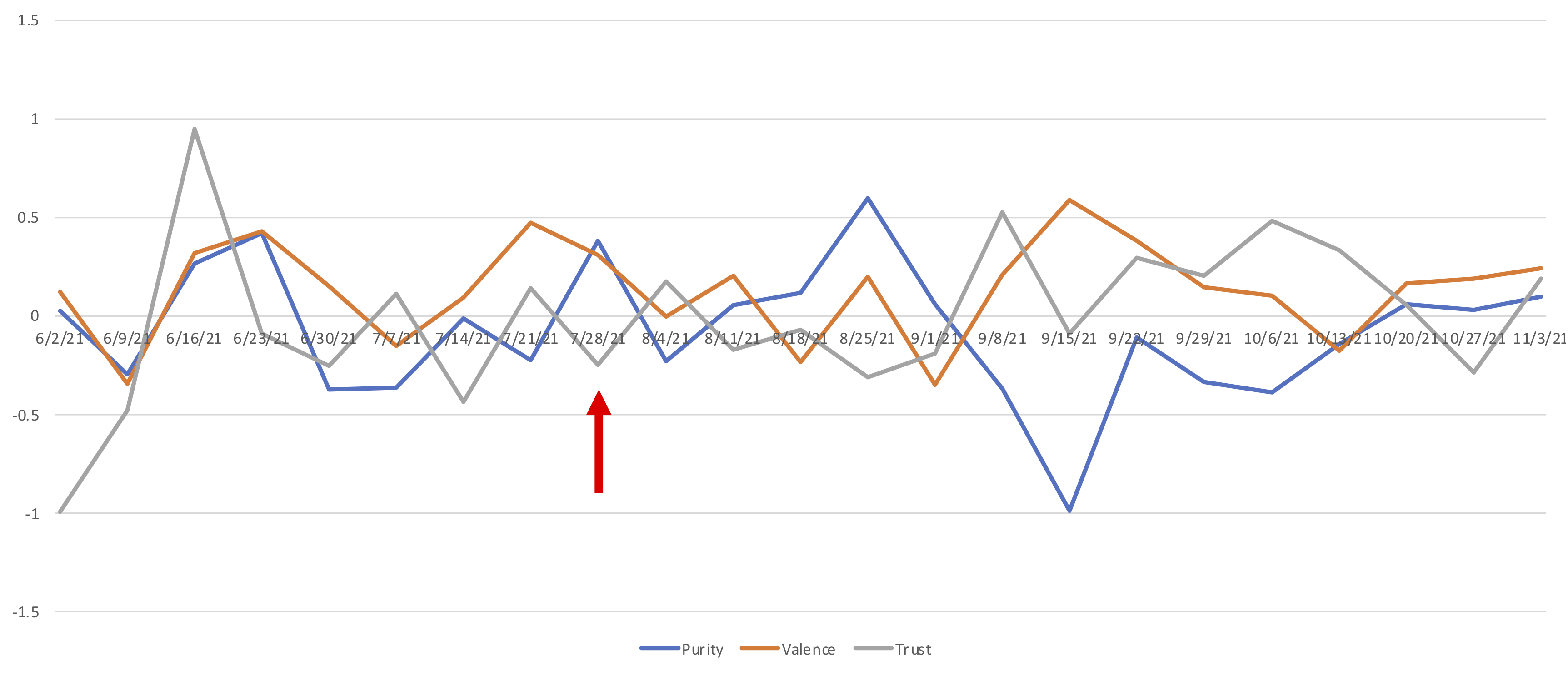}\hfill
\caption{Journalist F4's purity (blue), valence (orange), and trust (grey) scores over time. The red arrow corresponds to the July 22 tweet by Eduardo Bolsonaro.}
\label{fig:F4}
\end{figure}

On July 22, 2021, Eduardo Bolsonaro accused journalist F4 of corruption because her company allegedly received a contract from the federal government that was brokered by her ex-husband, a Brazilian senator. Although these events seemed to correspond to a slight decrease in journalist F4's trust score, this decrease was small in magnitude, perhaps due to counterspeech for journalist F4's supporters (Figure \ref{fig:F4}). 

\subsubsection{Journalist M1}
\begin{figure}[htp]

\centering
\includegraphics[width=.45\textwidth]{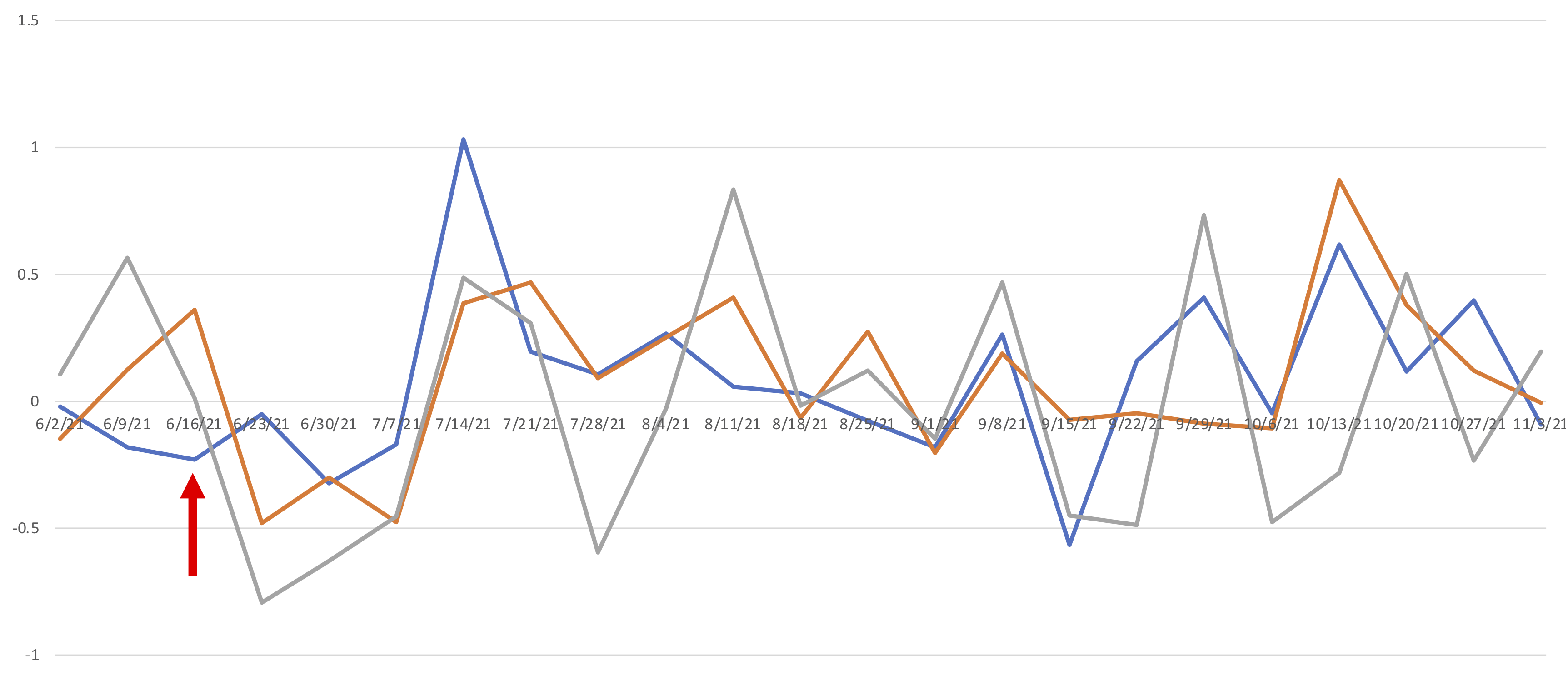}\hfill
\caption{Journalist M1's purity (blue), valence (orange), and trust (grey) scores over time. The red arrow corresponds to the July 22 tweet by the Special Advisor to the President.}
\label{fig:M1}
\end{figure}

On June 12, 2021, the Special Advisor to the President attacked journalist M1 on Twitter, accusing him of having fabricated the ``hate cabinet'' concept. This attack occurred in the middle of the week, and overlapped with a slight reduction in journalist M1's purity score, but was followed by a decline in his trust score in the following week (Figure \ref{fig:M1}).

\subsubsection{Journalist M2}
Journalist M2 was attacked four times on Twiter -- twice by Eduardo Bolsonaro (on July 22 and August 20, 2021), and twice by Carolos Bolsonaro (On August 21 and September 21, 2021). Despite the frequency of these attacks, there does not appear to be a consistent relationship between the dates of these attacks at journalist M2's scores.

\begin{figure}[htp]

\centering
\includegraphics[width=.45\textwidth]{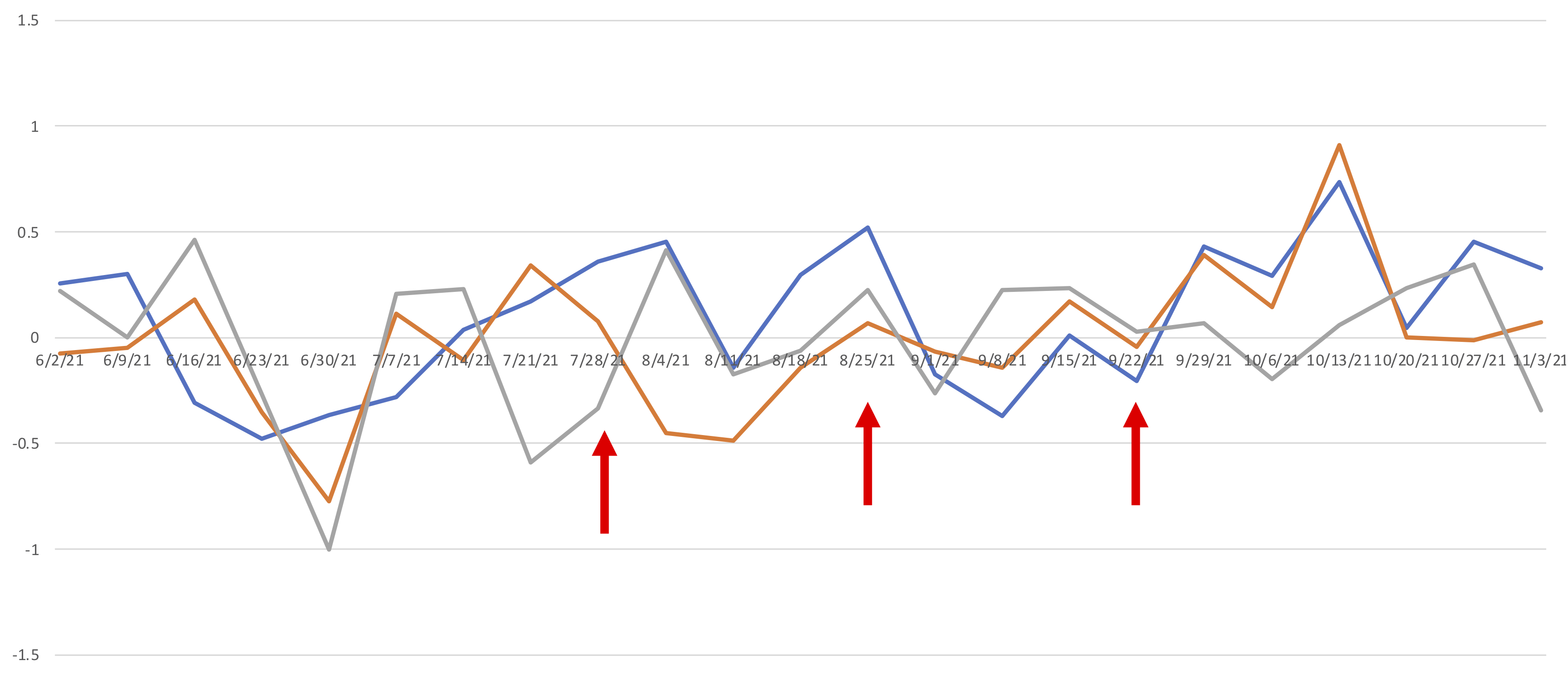}\hfill
\caption{Journalist M2's purity (blue), valence (orange), and trust (grey) scores over time. The red arrows correspond to tweets by the Bolsonaro brothers.}
\label{fig:M2}

\end{figure}
\subsubsection{Journalist M3}
\begin{figure}[htp]

\centering
\includegraphics[width=.45\textwidth]{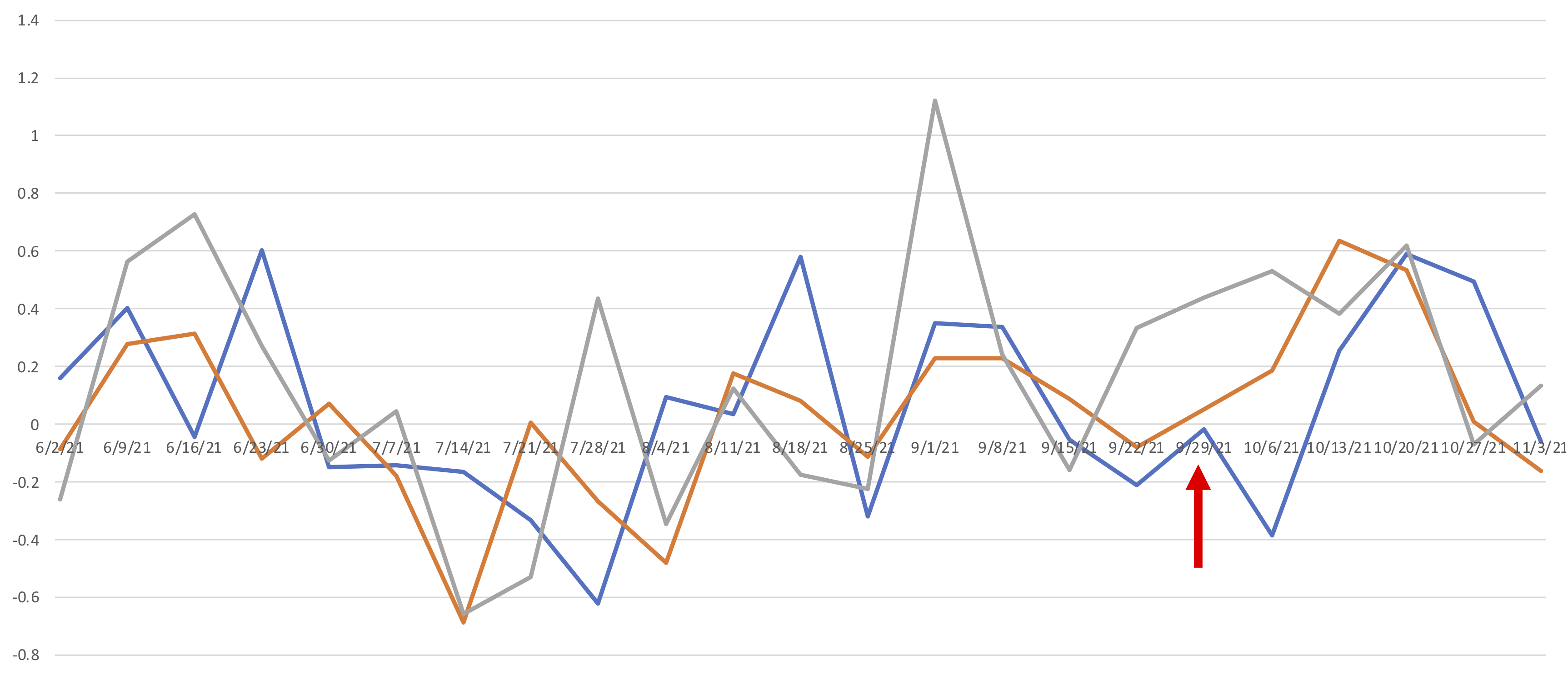}\hfill
\caption{Journalist M3's purity (blue), valence (orange), and trust (grey) scores over time. The red arrows correspond to tweets by Carlos Bolsonaro.}
\label{fig:M3}

\end{figure}

On September 24, 2021, journalist M3 criticized the government for inadequate COVID-19 precautions, including limited mask wearing, limited social distancing, and COVID denialism. Carlos Bolsonaro responded the next day (September 25), with a tweet crude tweet containing sexual innuendo. Journalist M3's purity score declined in the following week (Figure \ref{fig:M3}).

\section{Discussion}
Our method replicates the success of prior work \cite{toney2021automatically} demonstrating that information operations leave traces that may be detected using techniques designed to evaluate the overall biases in large corpora \cite{caliskan2017semantics}. Beyond this prior work, our analysis demonstrates the gendered nature of attacks targeting journalists in Brazil. Although reported in the popular press, we are the first to provide a systematic account of these attacks and the specific ways in which they appear designed to undermine specific journalists' reputation. In particular, women are framed as both less pleasant, less trustworthy, and less ``pure'' -- i.e., more disgusting and immoral -- than male journalists. 
To our knowledge, this work is also the first to use psychometric methods, and especially the MTMM, to evaluate the construct validity of our tools. Using this technique, we were able to demonstrate the convergent-discriminant validity of valence, trust, and purity scores. As expected, these three quantities are all significantly correlated with one another, but not correlated with random words. Notably, this technique also allows us to assess the performance of specific word embedding algorithms against others. In particular, we found the best performance (i.e., highest reliability and best convergent-discriminant validity) when using a model pre-trained using Word2Vec with the SGNS algorithm, but fine-tuned using the CBOW algorithm. We do not claim that this combination of algorithms is the best in all cases -- rather, we demosntrate a technique that future work could use to assess several different algorithms.

Althought we did not have a large enough sample of attacks against specific journalists by political figures to conduct a statistically significant analysis, our diachronic analyses provides several qualitative insights into how trolling campaigns might the affect overall discourse surrounding a particular journalist. In two out of four cases, specific attacks targeting these journalists corresponded to visible decreases in their trust scores; however, different attacks appeared to correspond to different dynamics. For example, Journalist F3's attack appeared to happen after, rather than before, a decline in her trust score indicating that, in some cases, the specific mention by a political figure may be signal the culmination, rather that the commencement, of an attack. Whereas politicians may lead the crowd in some cases, they may sense the direction of the crowd and jump in front in others. Furthermore, our inability to detect even a visual difference for Journalist F4 indicates that our technique may require refinement and, in particular, may be sensitive to counterspeech. Finally, we note that male journalists appear to trigger different dynamics than female journalists, with reductions in trust, purity, and valence scores occurring in a delayed manner if at all in the cases of journalists M1, M2 and M3. 

In general, our results seem to indicate that reductions in trust, valence, and purity scores do not seem to last longer than 1-2 weeks after an attack, but longer-term, longitudinal data can analyze these trends to see if consistent waves of attacks might leave lasting traces. Future work can therefore use our measures in a quasi-experimental setting.

\section*{Acknowledgment}

We would like to acknowledge of João Guilherme Bastos dos Santos for his invaluable assistance.

\bibliographystyle{IEEEtran} 
\bibliography{references} 

\section*{Appendix A}
Abusive words from the \#GloboLixo corpus selected by domain experts:
\textit{asco, asquerosa, asqueroso, assaltante, babaca, bandida, bandido, baranga, bruaca, burra, burro, canalha, chorar, cínica, cínico, comunista, corrupta, corrupto, covarde, cuzão, cuzona, demônio, descarada, descarado, desgraçada, desgraçado, divulgador de fake news, doente, doida, doido, escória, escrota, escroto, espalhador de fake news, esquerdista, fanática, fanático, frescura, gado, guerrilheira ,guerrilheiro, hipócrita, idiota, imbecil, incapaz, jumenta, jumento, ladra, ladrão, lamentar, lixo, maldita, maldito, maricas, mediocre, merda, militante, nojenta, nojento, ordinária, ordinário, otária, otário, palhaça, palhaço, pateta, patética, patético, pilantra, propagador de fake news, rata, rato, retardada, retardado, safada,
safado, terrorista, velhaca, velhaco, verme}

\section*{Appendix B}
This appendix contains an extended MTMM, with entries on the diagonal constituting reliability scores (Cronbach's $\alpha$ values) across ten replications of each model and entries on the off-diagonals indicating Pearson correlations between average model scores for each journalist present in the \#GloboLixo dataset (n=58). Correlations with absolute values of 0.28 or larger are statistically significant at the $p=0.05$ level. We refer to each algorithm by the algorithm used to pre-train the corpus followed by the algorithm used for fine-tuning. For example, a corpus pre-trained using SGNS but fine-tuned using CBOW is SGNS/CBOW. Models are referred to as ``locked'' if the fine-tuning was not allowed to update the pre-trained vectors (i.e., using Gensim\cite{rehurek2011gensim}, the $lockf$ parameter was set to 0. For unlocked models, this parameter was set to 1.0.
\FloatBarrier
\begin{table}[h]
\begin{tabular}{lllccccc}
1  & Locked SGNS/CBOW   & Purity    & 0.95  &       &       &       &       \\
2  &                    & Valence   & 0.51  & 0.93  &       &       &       \\
3  &                    & Random    & 0.1   & 0.03  & 0.96  &       &       \\
4  &                    & Trust     & 0.56  & 0.34  & 0.21  & 0.92  &       \\
5  &                    & Saturated & 0.66  & 0.4   & 0.08  & 0.46  & 0.92  \\
6  & Locked SGNS/SGNS   & Purity    & 0.62  & 0.3   & -0.11 & 0.34  & 0.25  \\
7  &                    & Valence   & 0.09  & 0.41  & -0.12 & -0.09 & -0.05 \\
8  &                    & Random    & 0.14  & 0.09  & 0.44  & 0.14  & -0.02 \\
9  &                    & Trust     & 0.29  & -0.15 & 0.16  & 0.53  & 0.08  \\
10 &                    & Saturated & 0.4   & 0.24  & 0.15  & 0.41  & 0.35  \\
11 & Unlocked CBOW/SGNS & Purity    & 0.28  & 0.1   & -0.24 & 0.09  & 0.24  \\
12 &                    & Valence   & 0.14  & 0.45  & -0.08 & 0.03  & 0.17  \\
13 &                    & Random    & 0.08  & 0.02  & 0.11  & -0.06 & 0.11  \\
14 &                    & Trust     & 0.08  & -0.19 & 0.12  & 0.31  & 0.17  \\
15 &                    & Saturated & 0.24  & 0.09  & 0.08  & 0.43  & 0.5   \\
16 & Locked GloVe/SGNS  & Purity    & 0.18  & -0.25 & 0.21  & 0.04  & 0.22  \\
17 &                    & Valence   & -0.64 & -0.47 & 0.01  & -0.26 & -0.31 \\
18 &                    & Random    & 0.24  & 0.25  & -0.05 & 0.02  & -0.07 \\
19 &                    & Trust     & 0.25  & 0.18  & 0.28  & 0.15  & 0.13  \\
20 &                    & Saturated & 0.03  & 0.05  & 0.2   & -0.05 & -0.06
\end{tabular}
\caption{\label{tab:subtable1}MTMM showing correlations between all models and the Locked Word2Vec SNGS model with CBOW fine-tuning.}
\end{table}
\FloatBarrier
\begin{table}[h]
\begin{tabular}{lllccccc}
   &                    &           & 6     & 7     & 8     & 9     & 10                   \\
6  & Locked SGNS/SGNS   & Purity    & 0.87  &       &       &       &                      \\
7  &                    & Valence   & 0.23  & 0.84  &       &       &                      \\
8  &                    & Random    & 0.25  & 0.05  & 0.84  &       &                      \\
9  &                    & Trust     & 0.3   & -0.3  & 0.29  & 0.84  & \multicolumn{1}{l}{} \\
10 &                    & Saturated & 0.65  & 0.23  & 0.11  & 0.21  & 0.83                 \\
11 & Unlocked CBOW/SGNS & Purity    & 0.38  & 0.1   & -0.19 & 0.03  & 0.34                 \\
12 &                    & Valence   & 0.1   & 0.66  & 0.04  & -0.21 & 0.2                  \\
13 &                    & Random    & 0.13  & 0.27  & 0.41  & 0.14  & 0.21                 \\
14 &                    & Trust     & 0.06  & -0.26 & -0.04 & 0.67  & 0.16                 \\
15 &                    & Saturated & 0.24  & 0.12  & 0.04  & 0.12  & 0.57                 \\
16 & Locked GloVe/SGNS  & Purity    & -0.04 & -0.15 & -0.04 & 0.05  & -0.06                \\
17 &                    & Valence   & -0.53 & -0.19 & 0.09  & 0     & -0.43                \\
18 &                    & Random    & 0.28  & 0.18  & 0.12  & 0.13  & 0.1                  \\
19 &                    & Trust     & -0.11 & -0.02 & 0.14  & 0.03  & -0.19                \\
20 &                    & Saturated & 0.09  & 0.09  & 0.15  & 0.08  & 0.02                
\end{tabular}
\caption{\label{tab:subtable2}MTMM showing correlations between all models and the Locked Word2Vec SNGS model with SGNS fine-tuning.}
\end{table}
\FloatBarrier
\begin{table}[h]
\begin{tabular}{lllccccc}
   &                    &           & 11    & 12   & 13   & 14   & 15    \\
11 & Unlocked CBOW/SGNS & Purity    & 0.95  &      &      &      &       \\
12 &                    & Valence   & 0.48  & 0.92 &      &      &       \\
13 &                    & Random    & 0.44  & 0.57 & 0.88 &      &       \\
14 &                    & Trust     & 0.44  & 0.17 & 0.37 & 0.85 &       \\
15 &                    & Saturated & 0.17  & 0.2  & 0.21 & 0.24 & 0.89  \\
16 & Locked GloVe/SGNS  & Purity    & 0.16  & 0.00    & 0.16 & 0.17 & 0.17  \\
17 &                    & Valence   & -0.28 & -0.1 & 0.02 & 0.23 & -0.11 \\
18 &                    & Random    & 0.26  & 0.32 & 0.37 & 0.08 & 0.01  \\
19 &                    & Trust     & 0.02  & 0.12 & 0.11 & 0.05 & -0.04 \\
20 &                    & Saturated & 0.27  & 0.28 & 0.33 & 0.29 & 0.03 
\end{tabular}
\caption{\label{tab:subtable3}MTMM showing correlations between all models and the Unlocked Word2Vec CBOW model with SGNS fine-tuning.}
\end{table}
\FloatBarrier
\begin{table}[h]
\begin{tabular}{lllccccc}
   &                   &           & 16   & 17   & 18   & 19   & 20   \\
16 & Locked GloVe/SGNS & Purity    & 0.92 &      &      &      &      \\
17 &                   & Valence   & 0.18 & 0.92 &      &      &      \\
18 &                   & Random    & 0.11 & -0.2 & 0.94 &      &      \\
19 &                   & Trust     & 0.52 & 0.1  & 0.27 & 0.86 &      \\
20 &                   & Saturated & 0.44 & 0.13 & 0.51 & 0.57 & 0.86
\end{tabular}
\caption{\label{tab:subtable4}MTMM showing correlations between all models and the Locked GloVe model with SGNS fine-tuning.}
\end{table}
\FloatBarrier
\end{document}